\documentclass{article} % For LaTeX2e
\usepackage{PRIMEarxiv}

\usepackage{amsmath,amsfonts,bm}

\def\eqref#1{equation~\ref{#1}}
\def\1{\bm{1}}

\DeclareMathAlphabet{\mathsfit}{\encodingdefault}{\sfdefault}{m}{sl}
\SetMathAlphabet{\mathsfit}{bold}{\encodingdefault}{\sfdefault}{bx}{n}

\usepackage{hyperref}
\usepackage{url}

\usepackage[T1]{fontenc}    % use 8-bit T1 fonts
\usepackage{hyperref}       % hyperlinks
\usepackage{url}            % simple URL typesetting
\usepackage{booktabs}       % professional-quality tables
\usepackage{amsfonts}       % blackboard math symbols
\usepackage{nicefrac}       % compact symbols for 1/2, etc.
\usepackage{microtype}      % microtypography
\usepackage{sidecap}
\usepackage{microtype}
\usepackage{graphicx}
\usepackage{dblfloatfix} 
\usepackage{amssymb}
\usepackage{enumitem,kantlipsum}
\usepackage[table,x11names]{xcolor}
\usepackage{soul}
\usepackage{booktabs} % for professional tables
\usepackage[table]{xcolor}
\usepackage{graphicx}
\usepackage{dblfloatfix} 
\newcommand\Alpha{\mathrm{A}}
\newcommand\Beta{\mathrm{B}}
\usepackage{xcolor}
\let\oldnl\nl% Store \nl in \oldnl
\newcommand{\nonl}{\renewcommand{\nl}{\let\nl\oldnl}}% Remove line number for one line

\usepackage{adjustbox}
\usepackage{setspace}
\usepackage[ruled, lined, linesnumbered, commentsnumbered, longend]{algorithm2e}

\SetCommentSty{mycommfont}

\usepackage{anyfontsize}
\usepackage{dblfloatfix}   
\usepackage{enumitem,kantlipsum}

\usepackage{floatrow}
\usepackage{tabularx}
\usepackage{amsfonts}
\usepackage{amssymb}
\usepackage{amsmath}
\usepackage{caption}

\usepackage{dsfont}
\usepackage{mathtools}
\usepackage{amsthm}
\usepackage{multirow}
\usepackage[normalem]{ulem}
\useunder{\uline}{\ul}{}
\usepackage{multirow}

\usepackage{microtype}
\usepackage{graphicx}
\usepackage{booktabs} % for professional tables

\title{Weakly-supervised Representation Learning for Video Alignment and Analysis}

\author{Guy Bar-Shalom$^*$ \\
Verily Research \\
guy.b@cs.technion.ac.il \\
\And
George Leifman \\
Verily Research \\
gleifman@verily.com \\
\And
Michael Elad \\
Verily Research \\
melad@verily.com \\
\And
Ehud Rivlin \\
Verily Research \\
ehud@verily.com \\
}
\begin{document}
% \nolinenumbers

\maketitle
\def\thefootnote{*}\footnotetext{This project was performed during an internship of the first author in Verily.}

\begin{abstract}
Many tasks in video analysis and understanding boil down to the need for frame-based feature learning, aiming to encapsulate the relevant visual content so as to enable simpler and easier subsequent processing. While supervised strategies for this learning task can be envisioned, self and weakly-supervised alternatives are preferred due to the difficulties in getting labeled data. This paper introduces LRProp -- a novel weakly-supervised representation learning approach, with an emphasis on the application of temporal alignment between pairs of videos of the same action category. The proposed approach uses a transformer encoder for extracting frame-level features, and employs the DTW algorithm within the training iterations in order to identify the alignment path between video pairs. Through a process referred to as ``pair-wise position propagation'', the probability distributions of these correspondences per location are matched with the similarity of the frame-level features via KL-divergence minimization. The proposed algorithm uses also a regularized SoftDTW loss for better tuning the learned features. Our novel representation learning paradigm consistently outperforms the state of the art on temporal alignment tasks, establishing a new performance bar over several downstream video analysis applications.

\end{abstract}
\section{Introduction}

As in many other domains, deep learning techniques have brought a revolution to the field of video analysis and understanding in the past several years~\cite{kim2019self, wang2020deep, khan2022grounded}. Applications such as video classification~\cite{hao2022attention, zeng2019graph, wang2020attentionnas}, action detection~\cite{lei2021detecting, zhai20223d, han2022twinlstm}, video captioning~\cite{pan2020spatio, qiao2022action}, forecasting~\cite{chang2022strpm, wu2021greedy}, and many others, all have been getting new and highly effective AI-based solutions with unprecedented performance. Interestingly, within this impressive progress, the task of temporal alignment of video pairs has received relatively little attention. This paper offers a novel weakly-supervised approach towards representation learning for video, focusing on the temporal alignment application. 

% This paper focuses on this application and offers a novel self-supervised approach towards its solution via representation learning. 

%There are many examples of events happening in a specific temporal order,
Many events occur in a specific temporal sequence, such as specific actions in sport activity (e.g., baseball swing),
% plant growing from a seedling to a tree
portions of a person's daily routine, various repetitive medical procedures, sea tides, intervals within traffic control videos, or even simple actions such as pouring a glass of fluid. In all these and other cases, videos that capture such visual content contain not only information about the cause and effect of these events, but also the potential for temporal correspondences across multiple instances of the same process. For example, the key moment of reaching for a container and lifting it off the ground are common to all pouring sequences, despite differences in visual factors such as %viewpoint, scale, 
container style, illumination, surrounding items, and event speed. Broadly speaking, a reliable alignment of such videos can open the door to new abilities in video analysis, such as detecting anomaly behavior, enabling quick search of specific sub-events, identifying the phase within the whole event, measuring distances between videos for their clustering, and so much more. 

So, how can videos be aligned? Earlier work~\cite{misra2016shuffle, dwibedi2019temporal, chen2022frame} suggests to address this task by first learning spatio-temporal feature representations for each video frame. Given such representations, their sequential matching over time, which takes into account temporal correspondences, would result in the desired alignment~\cite{dwibedi2019temporal, purushwalkam2020aligning, haresh2021learning}. 
This temporal matching could be achieved in a variety of ways, ranging from the simple nearest neighbor search, all the way to Dynamic Time Warping, DTW~\cite{berndt1994using}. 
The sought representations should capture key visual information while discarding irrelevant details, and do this while also providing a dimensionality reduction for ease of later processing. 

Representation learning could be achieved in a supervised fashion when frame-by-frame alignment is readily available. In this case, our task simply involves learning a common embedding space from pairs of aligned frames. A few approaches have been proposed for such supervised action recognition and segmentation \cite{tran2015learning, carreira2017quo, farha2019ms}. However, in many real-world sequences, such frame-by-frame alignment does not exist naturally. Obtaining labels for every frame in a video might be time-consuming and ill-defined task, as it is unclear what set of labels would be necessary for a complete understanding of the fine-grained details of the video. A possible remedy to the above could be to artificially obtain aligned sequences by recording the same event from multiple cameras, but this method may not capture all the variations present in naturally occurring videos. All this brings us to self - or weak-supervision alternatives, as practiced in~\cite{chen2022frame, haresh2021learning}. These methods may rely on video augmentation or use a SoftDTW~\cite{cuturi2017soft} loss for training the representations.

\begin{figure}[h]
    \centering
    \includegraphics[width=8cm]{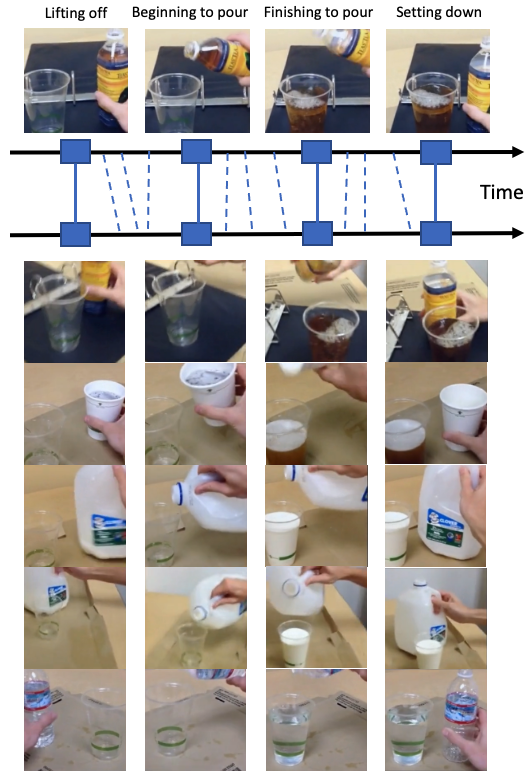}
    \caption{Video alignment (\textbf{Pouring} dataset) using \textbf{LRProp} features and the DTW algorithm. The first row shows a selected set of key events in a randomly selected video; the bottom shows the alignment results of  these events with five other randomly chosen videos. As can be seen, \textbf{LRProp} leads to a successful capture of the key events in the query. Note that we show a single time-line for all the five selected videos for simplicity.}
    \label{fig: alignment_two_column}
\end{figure}

Inspired by the above described work, we present {\bf LRProp} (\emph{Learning Representations by
position PROPagation}) -- a novel weakly-supervised approach that does not require explicit frame correspondences between different video sequences. 
To accomplish our goal, we adopt the transformer encoder~\cite{vaswani2017attention}, which has demonstrated effectiveness in extracting meaningful frame-level features, as shown in \cite{chen2022frame, arnab2021vivit, neimark2021video}. Our experiments suggest that the transformer encoder is particularly beneficial for video alignment, and we believe that this is due to its positional encoding. 
In each step within the learning process, our method involves taking pairs of videos that depict the same action category, feeding them into the transformer encoder to generate frame-level features, and using the DTW algorithm \cite{berndt1994using} to produce a path aligning the two. 
We use this alignment path to define a \emph{pair-wise position propagation} (definitions follow), which we utilize to establish a soft one-to-one linkage between pairs of frames. This \emph{pair-wise position propagation} is used to define a prior distribution for the correspondence between two distinct frames from different videos, and we minimize the KL-divergence between this prior and a probability distribution over the similarity of the frame-level features (extracted by the transformer encoder) of the specified pair of frames. Our learning also employs the SoftDTW algorithm \cite{cuturi2017soft} with appropriate regularization to prevent trivial solutions, and to learn a more accurate alignment path for pairs of videos. Figure~\ref{fig: alignment_two_column} demonstrates temporal alignment using the features extracted by the proposed method.

To summarize, our contributions include the following:

\begin{itemize}
    \item We present a general weakly-supervised framework for learning frame-wise representations with a focus on video alignment.
    
    \item The proposed \emph{pair-wise position propagation} %, which is a proper use of the alignment path calculated with the Dynamic Time Warping (DTW) algorithm between a pair of videos during training, 
    is shown to result in features that offer better temporal awareness compared to prior work.
    
    \item Our approach achieves superior performance to the state-of-the-art on various temporal understanding tasks on the Pouring~\cite{sermanet2018time} and PennAction~\cite{zhang2013actemes} datasets, setting a new performance benchmark for downstream tasks.
\end{itemize}

% In order to do that, we adopt the transformer encoder, which our experiments suggest that it provides an inductive bias for video alignment (because of the positional encoding). On top of that we use the well-known SoftDTW algorithm, with a proper regularization (to avoid trivial solutions), and in difference to previous works, we utilize the alignment path itself, extracted by DTW, to provide us with the soft one-to-one correspondance between frames of different videos.

% Some recent works [15, 41] have used cycle-consistency losses to align individual frames locally, while others have explored global alignment for video classification and segmentation [8, 5]. In this work, we adapt these global alignment ideas for video representation learning. There have been several approaches proposed for supervised action recognition [47, 7, 51, 48] and action segmentation [16, 34], but these require fine-grained annotations that can be costly [42]. Given the abundance of public video data and the high cost of fine-grained annotation, it is important to explore self-supervised methods. We are motivated by datasets and downstream tasks that specifically benefit from temporal alignment, such as video streams of semi-repetitive activities in manufacturing and surgery. In these cases, it is desirable to measure variability and anomalies [46, 24] across the datasets, and representations that optimize for temporal alignment may be particularly effective.

\section{Related Work}
\label{sec: related work}
% Self supervised is important.
% It is good in images.
% It is also good in videos, we focus on videos.
% we are more intrested in algment, transformer is important.

Since labeled data can be expensive and time-consuming to collect and annotate, and may not always be available in sufficient quantities for certain tasks, self-supervised learning has become a widely studied area in the field of deep learning. As a result, numerous pretext tasks have been proposed for image-based methods to achieve self-supervision. These include relative patch prediction \cite{doersch2015unsupervised}, jigsaw puzzle solving \cite{noroozi2016unsupervised}, colorization \cite{zhang2016colorful}, rotation prediction \cite{chen2019self}, instance discrimination using strong data augmentation \cite{chen2020simple}, knowledge distillation~\cite{chen2020big, caron2021emerging}, and more. These methods and many others have been shown to be highly effective in various downstream tasks. In this study, we investigate the use of self-supervised and weakly-supervised (definitions follow) learning techniques to create representations from videos, taking advantage of both the spatial and temporal information contained within the video data.

As we transition from images to video, various supporting tasks that produce supervision signals have been employed in representation learning for video frames.
% Various supervision tasks have been introduced as supervision signals for self-supervised learning from videos.
These include predicting the sequence of frames in a video~\cite{srivastava2015unsupervised, vondrick2016generating} or predicting audio from video~\cite{yadav2022learning}.
Recently, the incorporation of temporal ordering has been demonstrated to be a strong pretext task to obtain meaningful video representations. Sermanet et al.~\cite{sermanet2018time} proposed Time-Contrastive Networks (\textbf{TCN}), which uses attraction and repulsion between temporally close and far frames, respectively, in order to learn useful features. \textbf{Sal}~\cite{misra2016shuffle} proposed to learn features by sampling tuples of frames and predicting whether the tuple is in the correct temporal order (obtaining the labels using the frame indices). Another closely related work (\textbf{TCC}) was proposed by~\cite{dwibedi2019temporal}, which learns representations by finding frame correspondences across videos.

In this context, we should mention two recent works, \textbf{SCL} and \textbf{VAVA}~\cite{chen2022frame, liu2022learning}, which also achieved impressive results in self-supervised learning for videos. Chen et al.~\cite{chen2022frame} proposed strongly augmenting the input both temporally and spatially, and using a transformer model~\cite{vaswani2017attention} as an encoder, which has been used for videos in recent works~\cite{arnab2021vivit, neimark2021video}. They also used a contrastive loss function to encourage the embedding of nearby frames to be more similar than those that are far apart. Liu et al.~\cite{liu2022learning} attempted to learn video representations while considering the possibility of background frames, redundant frames, and non-monotonic frames when aligning two videos in time.

Whereas some work, as the above, assumes a self-supervised setting, %~\cite{haresh2021learning, chen2022frame}
in this paper we consider a weakly-supervised alternative, as described in~\cite{chang2019d3tw}. More specifically, we refer to cases in which the videos of interest consist of the same action category sequence. In these cases, we are given an ordered list of actions during training, but the exact temporal boundaries or paste of each action are not provided. For example, in a video of pouring wine, the weak supervision might include the sequence ``take the bottle, pour the wine, place the bottle back.'' This leads us to a powerful pretext task known as temporal video alignment, in which multitude of such corresponding videos can be leveraged for supervised learning. 

Though there is a significant amount of literature on time series alignment, only a few of these ideas have been applied to aligning videos. While traditional methods for time series alignment, such as DTW~\cite{berndt1994using}, are not differentiable and therefore cannot be used directly for training neural networks, a smooth approximation of DTW, called SoftDTW, was introduced in~\cite{cuturi2017soft}. Several recent papers~\cite{hadji2021representation, haresh2021learning} attempted to apply a soft DTW approximation in video representations learning.
As an example, we mention Haresh et al.~\cite{haresh2021learning}, which introduced a technique called \textbf{LAV} (Learning by Aligning Videos in Time) for learning representative frame embeddings by aligning videos in time using SoftDTW during training. However, their method, and other methods that use a soft DTW approximation during training, do not take advantage of the alignment path explicitly during optimization. In contrast, our approach, as unfolded in the next section, integrates both the SoftDTW cost function and the DTW alignment path into the optimization process.

\section{Method}
\label{sec: Method}
In this section we present \emph{Learning Representations by position PROPagation}, \textbf{LRProp}, a framework for learning frame-wise video representations. We learn an embedding space where videos with similar content can be aligned in time; this setting is commonly referred to as weakly-supervised learning, as discussed in Section~\ref{sec: related work}. More specifically, our method involves taking pairs of videos that depict the same action category, feeding them into the transformer encoder to generate frame-level features for each, and using the DTW algorithm \cite{berndt1994using} to produce a path aligning the two. We use this alignment to define a \emph{pair-wise position propagation}, and establish a soft one-to-one linkage between pairs of frames. We minimize the KL-divergence between a reference distribution and the probability function  over the similarity of the frame-level features of the specified pairs of frames. We also use the SoftDTW algorithm \cite{cuturi2017soft} with appropriate regularization to prevent trivial solutions as part of our training process. A visualization of our pair-wise position propagation method is depicted in Figure~\ref{fig: pair_wise_pos}. In what follows, we detail each of the above-described ingredients. 

\subsection{Notations and Definitions}
\label{sec: Notations and Definitions}

We begin by introducing the necessary notations and definitions for our discussion. Let $\mathcal{V}^1$ and $\mathcal{V}^2$ be a pair\footnote{While we discuss pairs of videos in each training step, our approach can be easily extended to any number of videos.} of videos, where each is represented as $\mathcal{V}^i \in \mathbf{R}^{F_i \times C \times W \times H}$. Here, $F_i$ represents the number of frames in the $i$th video and $C$, $W$, and $H$ are the number of channels, width, and height of each frame, respectively. To begin, we perform the same random sampling and data augmentation process described in \cite{chen2022frame}. This process takes a video $\mathcal{V}^i$ and uses temporal random cropping to generate two cropped videos of length $T$, $(V^i_1,S_1^i)$ and $(V^i_2,S^i_2)$, where $T$ is a hyper-parameter, and $S^i_{1/2}$ hold the frame indices. Next, several temporal-consistent spatial data augmentations are performed on each of the two sampled videos. After this step on $\mathcal{V}^1$ and $\mathcal{V}^2$, we are left with $(V^1_1, S^1_1), (V^1_2, S^1_2)$, and $(V^2_1, S^2_1), (V_2^2, S_2^2)$. 
We will use hereafter capital Latin letters to index the different videos, and greek letters to index the samplings.

% The variable $V^\Alpha_\alpha \in \mathbf{R}^{T \times C \times W \times H}$ represents the actual frames, and $S^\Alpha_\alpha \in \mathbf{R}^T$ represents the corresponding randomly chosen indexes of frames in the original video. Capital greek letters are used to index the videos, and regular greek letters are used to index the augmentations. For example, when we have two videos and two augmentations, we have $\Alpha \in { 1,2 }$ and $\alpha \in {1,2}$. 
% This preprocessing step is shown in Figure~\ref{fig:preprocessing}.

% \begin{figure*}[t]
%     \centering
% \includegraphics[width=\textwidth]{figures/method/preprocessing.jpg}
%     \caption{Preproccess.}
%     \label{fig:preprocessing}
% \end{figure*}

We define a neural model, $f_\theta: \mathbf{R}^{T \times C \times W \times H} \rightarrow \mathcal{Z}$, which maps videos from an input space to an embedding space $\mathcal{Z}$. We adapt the transformer model~\cite{vaswani2017attention} used by \cite{chen2022frame}. Assuming the existence of a prior distribution that represents the similarity between a frame, indexed $[S^\Alpha_\alpha]_i$, and a given frame, indexed $[S^\Beta_\beta]_j$, which we define by $p_{\Alpha,\Beta, \alpha, \beta}(i|j)$, our goal is to enforce the model's embedding ($f_\theta (V) \triangleq Z$) similarity to follow this distribution. We propose to achieve this by minimizing the KL-divergence between $p$ and the following distribution:
% given a frame indexes $[S^\Beta_\beta]_j$, the frame $[S^\Alpha_\alpha]_i$
% frame $[S^\Alpha_\alpha]_i$ will match frame $[S^\Beta_\beta]_j$, defined $p_{\Alpha,\Beta, \alpha, \beta}(i|j)$
% Our goal is to learn feature representations by aligning each of the $\binom{4}{2}$ videos in time. 
% More specifically, we use a prior distribution of the likelihood that frame $[S^\Alpha_\alpha]_i$ will match frame $[S^\Beta_\beta]_j$ only (assuming that frame $[S^\Beta_\beta]_j$ is the clear winner to match frame $[S^\Alpha_\alpha]_i$) defined as $p(i|j;\Alpha,\Beta, \alpha, \beta)$. We can enforce the model's embedding ($f_\theta (V) \triangleq Z$) similarity to follow this prior distribution by minimizing the KL-divergence between $p$ and the following distribution,
% We define the neural model, $f_\theta: \mathbf{R}^{T \times C \times W \times H} \rightarrow \mathcal{Z}$, to be a function of the input videos, to an embedding space, $\mathcal{Z}$. We wish to learn feature representations by aligning each of the $\binom{4}{2}$ videos in time.
% More rigorously, given a prior distribution of the likelihood of frame $[S^\Alpha_\alpha]_i$ to match frame $[S^\Beta_\beta]_j$ only (frame $[S^\Beta_\beta]_j$ is a clear winner to match frame $[S^\Alpha_\alpha]_i$), defined $p(i|j;\Alpha,\Beta, \alpha, \beta)$, we can enforce the model's embedding ($f_\theta (V) \triangleq Z$) similarity to follow the same prior distribution by minimizing the KL-divergence between $p$ and the following distribution, 
\begin{eqnarray}
\label{eq: embedding similarity}
 q_{\theta, \Alpha, \Beta, \alpha, \beta}(i|j) \equiv \mathcal{Q}_\theta (i|j) = 
 \frac{\text{exp}(\text{sim}([Z^\Beta_\beta]_j, [Z^\Alpha_\alpha]_i) / \tau) }
{\sum_{i'=1}^{T} \text{exp}(\text{sim}([Z^\Beta_{\beta}]_j,
[Z^\Alpha_\alpha]_{i^{'}})/\tau)}, 
\end{eqnarray}
where $\text{sim}$ denotes the cosine similarity ($\text{sim}(\mathbf{u},\mathbf{v}) = \mathbf{u}^T \mathbf{v}/ \lVert \mathbf{u} \rVert \lVert\mathbf{v} \rVert$) and $\tau$ is a hyper-parameter controlling the smoothness of the distribution.
The question remains, what would be a good prior for a given pair of videos?

%================================================

\subsection{Prior Distribution for Frames in the Same Video}
\label{sec: Prior Distribution for Two Augmented Copies of the Same Video}

% Regularization term
Let $(Z^\Alpha_\alpha, S^\Alpha_\alpha), (Z^\Beta_\beta, S^\Beta_\beta)$ be a given pair of embeddings, and their corresponding chosen indices. Following \cite{chen2022frame}, in the case where $\Alpha = \Beta$ (i.e., two sampled versions of the same video), the prior distribution is chosen as
\begin{eqnarray}
    \label{eq: prior for same video}
     p_{\Alpha,\Beta, \alpha, \beta}(i|j; \Alpha = \Beta) \equiv  \mathcal{P}_{\Alpha=\Beta}(i|j) = 
    \frac{\text{exp}(-([S^\Beta_\beta]_j - [S^\Alpha_\alpha]_i)^2 / 2\sigma^2) }
{\sum_{i'=1}^{T} \text{exp}(-([S^\Beta_\beta]_j - [S^\Alpha_\alpha]_{i^{'}})^2 / 2\sigma^2)}. 
\end{eqnarray}
% \begin{equation}
% \label{eq: prior for same video}
% p_{\Alpha, \Beta, \alpha, \beta}(i|j;\Alpha \neq \Beta) = \frac{\text{exp}(-([S^A_\alpha]_i - [S^B_\beta]_j)^2 / 2\sigma^2) }
% {\sum_{j'=1}^{T} \text{exp}(-([S^A_\alpha]_i - [S^B_\beta]_{j^{'}})^2 / 2\sigma^2)}
% \end{equation}
For the frame $j$ in video $\Beta$ sampled version $\beta$, this expression forms a  Gaussian of width $\sigma$ for nearby frames, centered around $[S^\Beta_\beta]_j$. 
Using this prior and optimizing the following objective,
\begin{equation}
    \label{eq: D_kl same}
    \mathcal{L}^j_{\Alpha = \Beta} = D_{KL}\big(\mathcal{P}_{\Alpha=\Beta}(\cdot|j) \parallel \mathcal{Q}_\theta(\cdot|j)\big),
\end{equation}
we enforce that the similarity between the embeddings of a given frame and its neighboring ones in the same video, follow such a Gaussian distribution. This assumption is reasonable, as adjacent frames are more correlated than far away ones. % Therefore, for a given frame $j$ in video $\Beta, \beta$, we can enforce the similarity of all frames in video $\Alpha, \alpha$ to follow the prior similarity distribution shown in Equation~(\ref{eq: prior for same video}) 
% of all frames $i$ in video $\Alpha, \alpha$ to frame this frame $j$, to follow the prior distribution in Equation~\ref{eq: prior for same video}, 
By accumulating the above over $j$, 
\begin{eqnarray}
\label{eq: loss for same videos}
    \mathcal{L}_{\text{Same}}(\mathcal{P}_{\Alpha=\Beta}, \mathcal{Q}_\theta) = \frac{1}{T} \sum_{j=1}^T \mathcal{L}^j_{\Alpha = \Beta},
\end{eqnarray}
we obtain the first loss component, which considers two different sampled versions of the same video. Observe that we omit the indices $\alpha, \beta$ for simplicity. 
%We optimize only in the case where $\alpha \neq \beta$, namely, $\alpha=1, \beta=2$, and $\alpha=2, \beta=1$. Since $\Alpha = \Beta$ and $\alpha = \beta$ correspond to the same video with the same augmentation.

%================================================

\subsection{Prior Distribution for Frames in Different Videos}
\label{sec: Prior Distribution for Two Augmented Copies of Two Different Videos}

In the case where $\Alpha \neq \Beta$, it is more challenging to define such a prior distribution. To match frames between different videos, we propose to rely on the path extracted by the Dynamic Time Warping (DTW) algorithm. To get a better understanding of our proposed method, we introduce the necessary notations and definitions regarding DTW. 

Given two sets of extracted embeddings, $Z^1 = \{ z^1_1, z^1_2, \ldots, z^1_n \}$ and $Z^2 = \{ z^2_1, z^2_2, \ldots, z^2_m \}$ from two input videos of lengths $n$ and $m$, respectively, we can compute the instantaneous distance matrix $D \in \mathbf{R}^{n \times m}$, where each entry is defined as $D(i,j) = d(z^1_i, z^2_j)$. The function $d: \mathcal{Z} \times \mathcal{Z} \rightarrow \mathbf{R}$ is a generic distance measure, implemented in this paper using the $l_2$-norm.
DTW computes the alignment loss between $Z^1$ and $Z^2$ by identifying the path of minimum cost in the distance matrix $D$:
\begin{equation}
    \text{\emph{dtw}}(Z_1, Z_2) = \text{\emph{min}}_{A \in \mathcal{A}_{n,m}} \langle A,D \rangle.
\end{equation}

The matrix $\mathcal{A}_{n,m} \subset \{ 0,1 \}^{n\times m}$ denotes the collection of all possible alignment matrices that correspond to paths from the top-left corner to the bottom-right corner of $D$, using only $\{ \rightarrow, \searrow, \downarrow \}$ moves. The alignment matrix $A \in \mathcal{A}_{n,m}$ is binary, where $A(i,j) = 1$ indicates that the embedding $z^1_i$ from $Z^1$ is aligned with the embedding $z^2_j$ from $Z^2$. 
DTW can be computed using dynamic programming by applying the following recursive function,
\begin{eqnarray}
    r(i,j) & = & D(i,j) +\text{min}\{ r(i-1,j), r(i,j-1), r(i-1,j-1) \}.
\end{eqnarray}
where $r(i,j)$ represents the (accumulated) DTW  distance between the two videos till their frames, $i$ and $j$, respectively. The minimum function is taken over all possible pairs of previous elements in the two sequences.

\begin{figure}[h]
    \centering
    \includegraphics[width=\linewidth]{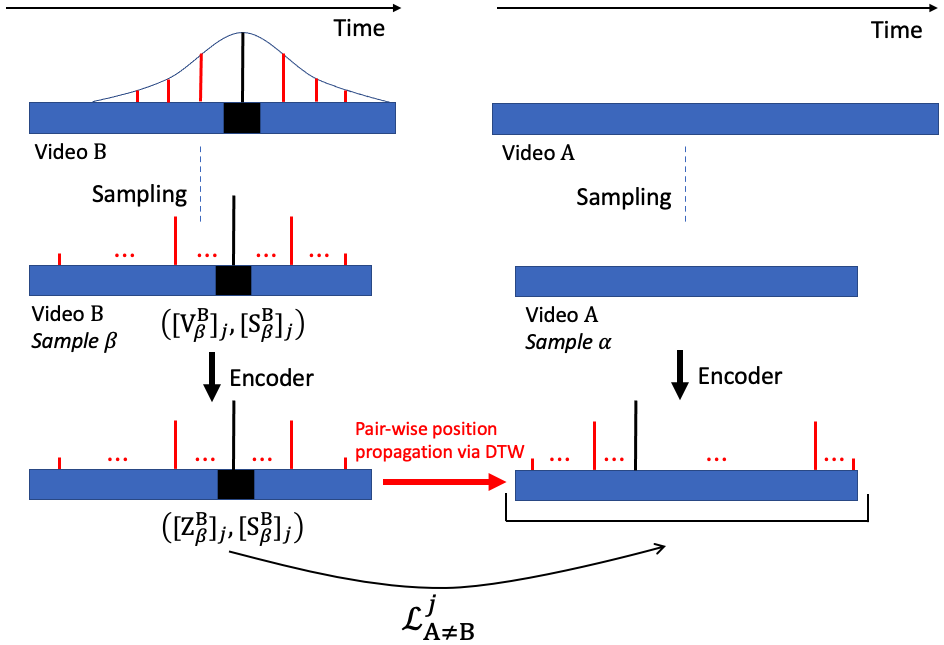}

        \caption{Illustration of pair-wise position propagation, showing the loss computation for the $j$th frame in the $\beta$ sample of video $\Beta$. We first calculate the Gaussian distribution of timestamp distances, centered around this frame. Afterwards, we propagate this distribution from video $\Beta$ to video $\Alpha$ and minimize the KL-divergence between that distribution and the embedding similarity one.}

    % \caption{We introduce \textbf{LRProp}, a weakly-supervised framework for learning video representations. 
    % \textbf{LRProp} begins by performing spatio-temporal sampling, and then optimizes the embedding space using the SoftDTW ($\mathcal{L}_{\text{Sdtw}}$) algorithm and enforcing a similarity prior distribution between frames from different videos through the use of two loss objectives, $\mathcal{L}_{\text{Same}}$ and $\mathcal{L}_{\text{Prop}}$. $\mathcal{L}_{\text{Same}}$ ensures that a frame is more similar to its neighboring frames than to distant ones, while $\mathcal{L}_{\text{Prop}}$ aims to achieve the same goal when dealing with frames from two different videos. To achieve this, \textbf{LRProp} propagates the positions of the frames, $\boldsymbol{S^1}$, from video 1 to video 2. The propagation is determined by the DTW alignment path. Our learned video representations can be useful for a variety of temporal understanding tasks, and are particularly effective for temporal video alignment.}
    \label{fig: pair_wise_pos}
\end{figure}

Returning to our video embedding task, we propose to use the alignment matrix $A$ to model the prior distribution for pairs of videos where $\Alpha \neq \Beta$. By generalizing the expression in Equation~(\ref{eq: prior for same video}), this prior is defined as follows:
% \begin{eqnarray}
% \label{eq: prior for different video}
% && p_{\Alpha,\Beta, \alpha, \beta}(i|j; \Alpha \neq \Beta) \equiv \mathcal{P}_{\Alpha \neq \Beta}(i|j) = \\
% && \frac{\text{exp}(-([S^B_\alpha]_j - [S^B_\alpha]_{\underset{k}{\mathrm{argmax}}\, A(k,i)})^2 / 2\sigma^2) }
% {\sum_{i^\prime=1}^{T} \text{exp}(-([S^B_\alpha]_j -[S^B_\alpha]_{\underset{k}{\mathrm{argmax}}\, A(k,i^\prime)})^2 / 2\sigma^2)} \nonumber .
% \end{eqnarray}
\begin{eqnarray}
\label{eq: prior for different video}
p_{\Alpha,\Beta, \alpha, \beta}(i|j; \Alpha \neq \Beta) \equiv \mathcal{P}_{\Alpha \neq \Beta}(i|j) = \frac{\text{exp}(-([S^\Beta_\beta]_j - [S^\Beta_\beta]_{\mathrm{argmax}_k\, A(k,i)})^2 / 2\sigma^2) }
{\sum_{i^\prime=1}^{T} \text{exp}(-([S^\Beta_\beta]_j -[S^\Beta_\beta]_{\mathrm{argmax}_k\, A(k,i^\prime)})^2 / 2\sigma^2)} .
\end{eqnarray}
Here, $i$ is a frame in video $\Alpha$, and $j$ is a frame in the video $\Beta$. The alignment matrix $A$ has rows corresponding to video $\Beta$ and columns corresponding to video $\Alpha$; therefore, ${\mathrm{argmax}}_k\, A(k,i)$ is the index of the frame in video $\Beta$ that has the maximum alignment score (which is $1$) with frame $i$ in video $\Alpha$. We define this process as \emph{pair-wise position propagation}. If there are multiple frames with the same maximum alignment score, we use the frame with the smallest index.
%We define Equation~(\ref{eq: prior for different video}) as the \emph{pair-wise position propagation} of frame $i$ in one video, to a given frame $j$ in a second video. 
We should emphasize that in the learning process this probability distribution is not differentiated with respect to the matrix $A$. Rather, this alignment matrix is updated during training (due to the modified representations), and considered as fixed when optimizing for the representations. 
%learned explicitly, but rather implicitly, because the alignment matrix is changing dynamically during training. 
Therefore, similarly to Equation~(\ref{eq: D_kl same}), given a frame $j$ in video $\Beta$, we propose to optimize
\begin{equation}
    \label{eq: D_kl diff}
    \mathcal{L}^j_{\Alpha \neq \Beta} = D_{KL}\big(\mathcal{P}_{\Alpha \neq \Beta}(\cdot|j) \parallel \mathcal{Q}_\theta(\cdot|j)\big),
\end{equation}
and accumulate all these divergence values as our second loss component,
\begin{eqnarray}
\label{eq: loss for different videos}
     \mathcal{L}_{\text{Prop.}}(\mathcal{P}_{\Alpha \neq \Beta}, \mathcal{Q}_\theta)  =
    \frac{1}{T} \sum_{j=1}^T \mathcal{L}^j_{\Alpha \neq \Beta}.
\end{eqnarray}
A visualization of the construction of Equation~(\ref{eq: D_kl diff}) is depicted in Figure~\ref{fig: pair_wise_pos}. 
%again, we omitted the indexes $\alpha, \beta$ for simplicity. 
% As mentioned in Section~\ref{sec: Prior Distribution for Two Augmented Copies of the Same Video}, we optimize for $\alpha=1, \beta=2$, and $\beta=1, \alpha=2$.
% % Recalling our setting, where we have two 
%  different videos, $(Z^\Alpha_\alpha, S^\Alpha_\alpha), (Z^\Beta_\beta, S^\Beta_\beta)$ (where $\Alpha \neq \Beta$), we propose to implement the prior distribution as follows,

% \begin{eqnarray}
% \label{eq: prior for different video}
% && p(i|j;\Alpha \neq \Beta, \alpha, \beta) = \\
% && \frac{\text{exp}(-([S^A_\alpha]_i - [S^A_\alpha]_{\underset{i^\prime}{\mathrm{argmax}}\, A(i^\prime,j)})^2 / 2\sigma^2) }
% {\sum_{j^\prime=1}^{T} \text{exp}(-([S^A_\alpha]_i -[S^A_\alpha]_{\underset{i^\prime}{\mathrm{argmax}}\, A(i^\prime,j^\prime)})^2 / 2\sigma^2)} \nonumber .
% \end{eqnarray}
% If $\underset{i^\prime}{\mathrm{argmax}}\, A(i^\prime,j)$ has more then one maximal values, we use the smallest index obtaining this value. 

%================================================

\subsection{Better Alignment via SoftDTW}
\label{sec: Optimizing SoftDTW to Enhance Better Alignment Features}

The pair-wise position propagation as described above is effective only when the features used are representative enough. To improve this alignment during training, we suggest to also optimize the smooth DTW (SoftDTW) distance~\cite{cuturi2017soft}. This is defined as follows,
\begin{eqnarray}
\label{eq: soft dtw}
    r^\gamma(i,j) = D(i,j) + \text{min}^\gamma\{ r^\gamma(i-1,j), r^\gamma(i,j-1), r^\gamma(i-1,j-1) \},
\end{eqnarray}
% Here, $D(i,j)$ is the $l_2$ distance between the features extracted from frames $i$ and $j$, and 
The term $r^\gamma(i,j)$ holds the soft-DTW distance up to frames $i$ and $j$ in the two videos, respectively. The expression $\text{min}^\gamma$ is a smooth (and therefore differentiable) version of the $\text{min}$ function, defined as:
% where $\text{min}^\gamma$ is a smooth (and therefore differentiable) version of DTW, defined,
\begin{equation}
    \text{min}^\gamma \{a_1, a_2, \ldots, a_n\} = -\gamma \log\sum_{i=1}^n e^{\frac{-a_i}{\gamma}}.
\end{equation}% Note that when gamma approaches zero we converge to the discrete min function.
Note that $\text{min}^\gamma$ converges to the discrete $\min$ operator as $\gamma$ approaches zero. Therefore, when $\gamma$ is near zero, the smooth DTW distance will produce results that are similar to those of the discrete DTW.

Armed with the SoftDTW, given a single pair of videos with their randomly chosen indices, $(Z^\Alpha_\alpha, S^\Alpha_\alpha, Z^\Beta_\beta, S^\Beta_\beta)$, we optimize the following objective function:
\begin{eqnarray}
\label{eq: final loss}
      \mathcal{L}_{\textbf{LRProp}}(Z^\Alpha_\alpha, S^\Alpha_\alpha, Z^\Beta_\beta, S^\Beta_\beta)=
    \delta_{\Alpha \Beta} \cdot \mathcal{L}_{\text{Same}} +
     (1 - \delta_{\Alpha \Beta})\cdot (\lambda_1 \cdot  \mathcal{L}_{\text{Prop.}} + \lambda_2 \cdot 
    \mathcal{L}_{\text{Sdtw}}).
    %  &(1 - \delta_{\Alpha \Beta})\cdot&  \mathcal{L}_{\text{Prop.}} + \nonumber \\
    %  &\delta_{\Alpha \Beta}\cdot&  \lambda_1 \cdot \mathcal{L}_{\text{Same}} + \nonumber \\
    % &(1 - \delta_{\Alpha \Beta}) \cdot & \lambda_2 \cdot 
% \mathcal{L}_{\text{Sdtw}}.
\end{eqnarray}
% \begin{eqnarray}
%     & \mathcal{L}&(Z^\Alpha_\alpha, S^\Alpha_\alpha, Z^\Beta_\beta, S^\Beta_\beta) =\nonumber \\
%     &\sum_j&\big((1 - \delta_{\Alpha \Beta}) D_{KL}(q_\theta (i|j;\Alpha \neq \Beta, \alpha, \beta) || p (i|j;\Alpha \neq \Beta, \alpha, \beta)) \nonumber \\
%     &+&  \lambda_1 \cdot \delta_{\Alpha \Beta} D_{KL}(q_\theta (i|j;\Alpha = \Beta, \alpha, \beta) || p (i|j;\Alpha = \Beta, \alpha, \beta)) \big) \nonumber \\
%     &+& \lambda_2 \cdot 
% \text{Sdtw}(Z^\Alpha_\alpha, Z^\Beta_\beta),     
% \end{eqnarray}
% $q_\theta (i|j;\Alpha \neq \Beta, \alpha, \beta)$ is the learned distribution for pairs of videos where $\Alpha \neq \Beta$, and $p(i|j;\Alpha \neq \Beta, \alpha, \beta)$ is the proposed prior distribution defined earlier.
Here, $\delta_{\Alpha \Beta}$ is the Kronecker delta, which is 1 if $\Alpha = \Beta$ and 0 otherwise. $\mathcal{L}_{\text{Sdtw}} \equiv \mathcal{L}_{\text{Sdtw}}(Z^\Alpha_\alpha, Z^\Beta_\beta)$ is the smooth DTW distance between the two videos, which is computed using the embedding vectors $Z^\Alpha_\alpha$ and $Z^\Beta_\beta$ via Equation~(\ref{eq: soft dtw}). $\lambda_1$ and $\lambda_2$ are hyper-parameters that control the relative importance of the different terms in this objective function. For further implementation details and hyper-parameters, see Appendix~\ref{app: Implementation Details}.

% \begin{figure*}[b]
%     \centering
%     \includegraphics[width=\textwidth]{figures/method/dtw_intuition_2.jpg}
%     \caption{Intuition for dtw.}
%     \label{fig:dtw intuition}
% \end{figure*}

\section{Empirical Study}
\label{sec: Empirical Study}
\begin{figure*}[h]
    \centering
    \includegraphics[width=\textwidth]{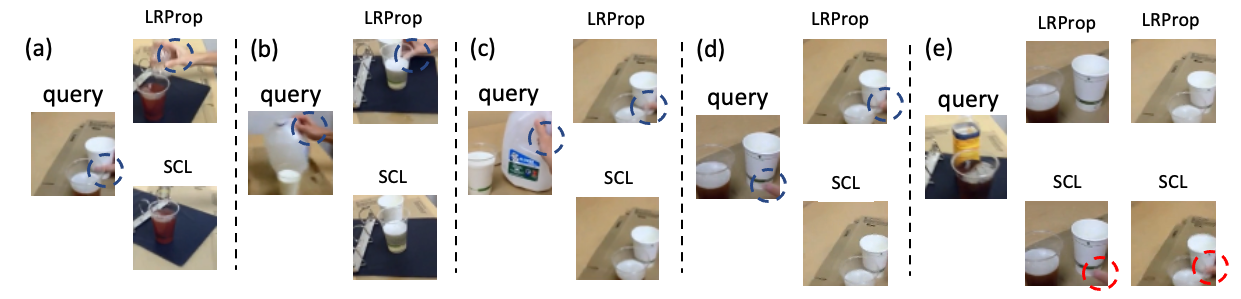}

    \caption{An example of aligned frames (\textbf{Pouring} dataset) using the DTW algorithm. In (a), (b), (c), and (d), \textbf{LRProp} is better at capturing the specific action of placing the bottle/cup, as indicated by the {\color{blue} blue} circles; in contrast, in SCL, the human hand is missing in the aligned frame. In (e), the query frame shows the end of the pouring action. Two aligned video results are shown; while \textbf{LRProp} provides a successful match, SCL features are inferior - observe the human hand still present in the frames, as indicated by the {\color{red} red} circles.}
    \label{fig: comparison}

\end{figure*}

In this section, we evaluate the performance of \textbf{LRProp} on two datasets using various evaluation metrics.

% We compare our method to the state-of-the-art technique, SCL \cite{xxx}, using the same datasets. We obtained the SCL code from their Github repository \footnote{https://github.com/minghchen/CARL\_code} and ran the code on our machine to compare the performance of the two methods. We add the following methods as baselines, XXX, which we reimplemented and replace the network \\

\subsection{Datasets}

The \textbf{PennAction dataset}~\cite{zhang2013actemes} includes videos of humans performing various sports and exercise activities. We use 13 of these actions, following TCC~\cite{dwibedi2019temporal}. The dataset includes a total of 1140 videos for training and 966 for testing, with each action set containing 40-134 train videos and 42-116 test videos. For evaluation, we obtain per-frame labels from LAV \cite{haresh2021learning}. These videos contain 18 to 663 frames. 

\textbf{Pouring dataset}~\cite{sermanet2018time}. This dataset contain videos showing the process of a hand pouring a liquid from one object to another. The phase labels, based on the TCC~\cite{dwibedi2019temporal}, consist of five classes. Following TCC, we use 70 videos for training and 14 for testing. These videos contain 186 to 797 frames. 

\subsection{Evaluation Metrics} 
We use the following metrics to evaluate the frame-wise trained representations: 

% \begin{itemize}
\noindent \textbf{\emph{Phase Classification Accuracy}}~\cite{dwibedi2019temporal} is a metric that measures the per-frame accuracy of Phase Classification. To calculate this metric, we first extract features from the frames in the training and test data. Then, we train a support vector machine 
(SVM)~\cite{hearst1998support} classifier on the phase labels for the training data, using the extracted features as input. The classifier is then used to predict the phase labels for each frame in the test data. The Phase Classification Accuracy is calculated as the proportion of correctly predicted labels, and is reported for different percentages of the SVM training set, in order to evaluate the representativeness of the learned features. 
\begin{table*}[t]
\small
\centering
\caption{Comparison with state-of-the-art methods on \textbf{Pouring} using various evaluation metrics: Phase Classification@\% (Classification@\%),
Phase Progression (Progress), Kendall’s Tau ($\tau$), Average Precision@K (AP@K), DTW Accuracy (DTW A). Best method is in \textbf{bold}, second best in {\ul underlined}. Our proposed technique (\textbf{LRProp}) % - highlighted in \sethlcolor{lightgray}\hl{gray})
dominates all other methods.}

\begin{tabular}{c|c|c|ccc|cccc|c}
\multirow{2}{*}{\textit{Method}} & \multirow{2}{*}{\textit{$\tau$}} & \multirow{2}{*}{\textit{Progress}} & \multicolumn{3}{c|}{\textit{AP@}}                             & \multicolumn{4}{c|}{\textit{Classification@}}        & \multirow{2}{*}{\textit{DTW A}} \\ 
                                  &                                         &                                             & \textit{K=5}    & \textit{K=10}   & \textit{K=15}   & \textit{10}     & \textit{25}     & \textit{50}     & \textit{100}   &                               \\ \midrule \midrule

SCL                               & {\ul 99.2}                              & {\ul 93.5}                                  & {\ul 90.04}$^\dagger$ & {\ul 89.69}$^\dagger$ & 88.92$^\dagger$ & 85.78$^\dagger$ & {\ul 87.14}$^\dagger$ & 89.45$^\dagger$ & 
 {\ul 93.73}          & {\ul 84.68}$^\dagger$               \\
SAL                               & 79.61                                   & 77.28                                       & 84.05           & 83.77           & 83.79           & 87.63           & -               & 87.58           & 88.81          & -                             \\
TCN                               & 85.12                                   & 80.44                                       & 83.56           & 83.31           & 83.01           & 89.67           & -               & 87.32           & 89.53          & -                             \\
TCC                               & 86.36                                   & 83.73                                       & 87.16           & 86.68           & 86.54           & 90.65           & -               & 91.11           & 91.53          & -                             \\
LAV                               & 85.61                                   & 80.54                                       & 89.13           & 89.13           & {\ul 89.22}           & 91.61           & -               & {\ul 92.82}           &  92.84          & -                             \\
VAVA                              & 87.55                                   & 83.61                                       & -               & -               & -               & {\ul 91.65}           & -               & 91.79           & 92.84         & -                       \\ \midrule % \rowcolor{lightgray}
\textbf{LRProp}                     & \textbf{99.46}                          & \textbf{94.09}                              & \textbf{92.41}  & \textbf{90.33}  & \textbf{90.86}  & \textbf{92.7}   & \textbf{93.88}  & \textbf{94.44}  & \textbf{94.36} & \textbf{90.22}                \\
\midrule \midrule     
\end{tabular}

\label{tab: main pouring}
\end{table*}

\noindent \textbf{\emph{Phase Progression}}~\cite{dwibedi2019temporal} evaluates the accuracy of the embeddings in representing the advancement of a process or action. To compute this metric, we establish a rough estimate of the progress within a phase by calculating the difference in timestamps between a specific frame and key events, and then normalizing this difference by the total number of frames in the video. This measure is used as the target for a linear regression model, which is trained on the frame-wise embeddings. The Phase Progression metric is then calculated as the average R-squared measure (coefficient of determination) of the regression model on the test data. This metric captures how well the embeddings capture the relative progression of the phase within a video.

\noindent \textbf{\emph{Average Precision@K}}~\cite{haresh2021learning} is a metric used to evaluate the accuracy of fine-grained frame retrieval, where K is the number of retrieved frames. To calculate this metric, we identify the K nearest frames (using KNN) to a given query frame, based on the frame-wise embeddings, and calculate the Average Precision, i.e., the average proportion of retrieved frames that have the same phase label as the query one. This metric is calculated for different values of K, to assess the performance of the frame-wise embeddings at different retrieval sizes. No additional training or fine-tuning is needed for calculating this metric.

\noindent \textbf{\emph{Kendall's Tau}}~\cite{dwibedi2019temporal} is a correlation coefficient that evaluates the temporal alignment of two sequences. Unlike the other metrics discussed above, Kendall's Tau does not require additional labels for evaluation. To calculate this metric, we first sample pairs of frames ($u_i, u_j$) from the first video, which has $n$ frames, and retrieve the corresponding nearest neighbors (in the feature space), ($v_p, v_q$), in the second video. This quadruplet of frame indices ($i, j, p, q$) is considered \emph{concordant} if $i < j$ and $p < q$ or $i > j$ and $p > q$. Otherwise, it is considered \emph{discordant}. Kendall's Tau is then calculated as the ratio of concordant pairs to the total number of pairs, which captures how well the two sequences are aligned in time. More specifically, it is defined over all pairs of frames in the first video as:
$\tau = (\text{no. of concordant pairs - no. of discordant pairs})/{n\choose 2}$.
Kendall's Tau is a measure of the alignment between two sequences in time, where a value of $1$ indicates perfect alignment and a value of $-1$ indicates that the sequences are aligned in the reverse order. One limitation of this metric is that it assumes that there are no repetitive frames in a video. We average this metric across all video pairs in the validation set.

\noindent\textbf{\emph{DTW Accuracy}} is the ultimate metric that measures the ability of the frame-wise embeddings to capture both the phase labels and the Phase Progression of actions or processes in videos. To calculate this metric, we first apply the dynamic time warping (DTW) algorithm to a pair of videos, using the frame-wise embeddings as input. This produces a sequence of connections between corresponding frames in the two videos. The DTW Accuracy is then calculated as the proportion of connections that connect frames with the same phase label. By definition, DTW does not allow for discordant indices,  % Therefore, the features that do not represent the phase progression well would result in previous or subsequent DTW connections, that do not agree on phase labels. Thus, 
and it's accuracy metric is sensitive to both the ability of the embeddings to predict the phase labels and the ability to capture the Phase Progression. Overall, DTW Accuracy is a useful metric for evaluating the effectiveness of frame-wise embeddings at capturing the temporal structure of actions or processes in videos. We evaluate this metric on all pairs of videos in the validation set, and take the average as the final result.

%--------------------
Following the work in \cite{haresh2021learning,dwibedi2019temporal, haresh2021learning, chen2022frame}, we use the first four metrics above to evaluate the effectiveness of frame-wise embeddings at capturing the temporal structure of actions or processes in videos. In addition, we propose to use the last metric (DTW Accuracy) in order to evaluate the suitability of the learned features for video alignment.
%, as it captures both the ability to predict phase labels and the ability to capture phase progression. This allows us to assess the performance of the frame-wise embeddings on a range of tasks that require accurate representation of the temporal structure of videos.

\subsection{Comparison to State-of-the-Art}
\begin{table*}
\small
\centering
\caption{Comparison with state-of-the-art methods on \textbf{PennAction} using various evaluation metrics. Best method is in \textbf{bold} and second best is {\ul underlined}. \textbf{LRProp} %(highlighted in \sethlcolor{lightgray}\hl{gray}) 
outperforms all previous methods.} 
% \caption{Comparison with state-of-the-art methods on \textbf{PennAction} using several metrics: Phase Classification@\% (Classification@\%),
% Phase Progression (Progress), Kendall’s Tau ($\tau$) and Average Precision@K (AP@K). Best method is in \textbf{bold} and second best is {\ul underlined}. \textbf{LRProp} highlighted in \sethlcolor{lightgray}\hl{gray}) outperforms all previous methods.} 

\begin{tabular}{c|c|c|ccc|cccc}
\multirow{2}{*}{\textit{Method}} & \multirow{2}{*}{\textit{$\tau$}} & \multirow{2}{*}{\textit{Progress}} & \multicolumn{3}{c|}{\textit{AP@}}                         & \multicolumn{4}{c}{\textit{Classification@}} \\
                                 &                                         &                                             & \textit{K=5}   & \textit{K=10} & \textit{K=15}  & \textit{10}        & \textit{50}         &   \textit{75} &\textit{100}       \\ \midrule \midrule

SCL                              & {\ul 98.5}                              & {\ul 91.8}                                  & {\ul 92.28}    & {\ul 92.1}    & {\ul 91.82}    & -                  & -              & -      & {\ul 93.07}        \\
SAL                              & 76.12                                    & 69.6                                      & -              & -             & -              & 74.87              & 78.26           & -      & 79.96            \\
TCN                              & 81.2                                    & 72.17                                       & 77.84          & 77.51         & 77.28          & 81.99              & 83.67          & -      & 84.04              \\
TCC                              & 81.35                                   & 73.53                                       & 76.74          & 76.27         & 75.88          & 81.26              & 83.35        & -         & 84.45             \\
LAV                              & 80.5                                    & 66.13                                       & 79.13          & 79.98         & 78.9           & {\ul 83.56}        & 83.95         & -       & 84.25              \\
VAVA                             & 80.53                                   & 70.91                                       & -              & -             & -              & 83.89              & {\ul 84.23}    & -      & 84.48    
 \\ \midrule % \rowcolor{lightgray}
 \textbf{LRProp}                    & \textbf{99.09}                          & \textbf{93.03}                              & \textbf{92.46} & \textbf{92.2} & \textbf{92.03} & \textbf{91.9}      & \textbf{92.96}   & \textbf{93.17}   & \textbf{93.25}     \\ \midrule \midrule

\end{tabular}

%Our proposed technique (\textbf{LRProp} - highlighted in \sethlcolor{lightgray}\hl{gray}) outperforms all the previous methods.}
\label{tab: main pen action}

\end{table*}

\textbf{Pouring Dataset.} In Table~\ref{tab: main pouring} we compare our method with state-of-the-art methods on the task of Pouring. The best method for each metric is shown in bold, and the second best is underlined. Our proposed method outperforms all previous work on this dataset, with SCL performing the second best overall. As not all of our proposed metrics were reported in their original paper, we reproduced their results using their Github repository\footnote{https://github.com/minghchen/CARL\_code.} for a fair comparison with our results -- these are marked by a $^\dagger$. 

Our method demonstrates superiority over all other methods in all tasks. In particular, with our approach we achieve a Phase Classification Accuracy of $93.88$\% by using only $25\%$ of the available labels, surpassing all other methods, even if they use 100\% of the labels. Additionally, our method excels at identifying frames with similar semantics from other videos, as demonstrated by an improvement of almost $2.5\%$ in the Average Precision@K (AP@) column. We also see significant gains in Kendall's tau and Phase progression metrics compared to SCL, which has already shown a phenomenal improvement of more than $10\%$ over all previous methods. Finally, the DTW Accuracy metric, which measures both Phase Classification and Phase Progression, shows a striking improvement of almost $6\%$ over the state-of-the-art, indicating that our proposed approach is highly effective for video alignment. This is also supported by Figure~\ref{fig: comparison}, which demonstrates the superior performance of \textbf{LRProp} compared to SCL in the alignment task: when given a query frame, \textbf{LRProp} is able to capture fine-grained actions more effectively. For further demonstration of \textbf{LRProp} on video alignment, see Figure~\ref{fig: alignment_two_column}, and the supplementary material.

\begin{table*}[b]
\tiny
\centering

\caption{An ablation study of $\mathcal{L}_{\text{Prop.}}$ and $\mathcal{L}_{\text{Sdtw}}$ using various evaluation metrics. Best result in each column is in \textbf{bold}. Our suggested method is in the bottom row.} %highlighted in \sethlcolor{lightgray}\hl{gray}.}

\begin{tabular}{llc||c|c|llllll|llllll|l}

\multirow{2}{*}{\textit{$\mathcal{L}_{\text{Same}}$}} & \multirow{2}{*}{\textit{$\mathcal{L}_{\text{Prop.}}$}} & \multirow{2}{*}{$\mathcal{L}_{\text{Sdtw}}$} & \multirow{2}{*}{\textit{$\tau$}} & \multirow{2}{*}{\textit{Progress}} &                \multicolumn{6}{c|}{\textit{AP@}}                                                                                                                                     & \multicolumn{6}{c|}{\textit{Classification@}}                                                                                                                                 & \multicolumn{1}{c}{\multirow{2}{*}{\textit{DTW A}}} \\ 
                                                      &                                                       &                                &                                                   &                                    & K=1            & \multicolumn{1}{c}{\textit{K=5}} & \multicolumn{1}{c}{\textit{K=10}} & \multicolumn{1}{c}{\textit{K=15}} & \textit{K=20}           & \textit{K=25}                     & \textit{5}             & \multicolumn{1}{c}{\textit{10}} & \multicolumn{1}{c}{\textit{25}} & \multicolumn{1}{c}{\textit{50}} & \textit{75}             & \multicolumn{1}{c|}{\textit{100}} & \multicolumn{1}{c}{}                      \\ \midrule \midrule
\checkmark                                                     & \checkmark                                                     &                               & 98.97                                             & 92.58                              & 92.74          & 90.4                             & 88.68                             & 87.49                             & 86.64          & 85.94                  & 81.99         & 83.99                           & 85.08                           & 87                              & 86.62          & 86.42                            & 81.38                                      \\
\checkmark                                                     &                                                      & \checkmark                              & \textbf{99.52}                                    & 92.25                              & 93.45          & 90.38                            & 89.02                             & 88.26                             & 87.55          & 86.69               & 86.66         & 87.66                           & 88.61                           & 89.75                           & 89.69          & 89.69                            & 85                                         \\ \midrule % \rowcolor{lightgray}
\checkmark                                                     & \checkmark                                                     & \checkmark                              & 99.46                                             & \textbf{94.09}                     & \textbf{94.09} & \textbf{92.41}                   & \textbf{91.66}                    & \textbf{90.86}                    & \textbf{90.45} & \textbf{90.07} & \textbf{91.8} & \textbf{92.7}                   & \textbf{93.88}                  & \textbf{94.44}                  & \textbf{94.46} & \textbf{94.36}                   & \textbf{90.22}       \\ \midrule \midrule                      

\end{tabular}

\label{tab: ablation}
\end{table*}

% \usepackage{multirow}
% \usepackage[normalem]{ulem}

% In Table~\ref{tab: main pen action}, we compare the performance of our method with state-of-the-art approaches on the PennAction dataset. Since our method is weakly supervised and specialized for alignment, we follow the approach of \cite{dwibedi2019temporal} and train a separate model for each of the 13 action classes in PennAction. The results shown in the table are the average across all 13 actions (For results for each class individually, see Appendix~\ref{xxx}). The best method for each metric is highlighted in bold, and the second best is underlined. As shown in the table, our proposed method consistently outperforms all prior work on the PennAction dataset. SCL performs the second best overall, but it should be noted that SCL trained a single model for all 13 action classes (they did not report results for each class individually). We further notice, that we require only 75\% labels to surpass all baselines in Phase classification. We also achieve more than 1\% improvement in the phase progression metric, which suggests that our method produces features that are more reliable for video alignment, or fine-grained retrieval. This is also supported by the small but consistent imporvement in the Kendell's tau metric, and the Average Precision @ K metric.

\textbf{PennAction Dataset.} As shown in Table~\ref{tab: main pen action}, our proposed method demonstrates superior performance in comparison to all other state-of-the-art approaches on the PennAction dataset. Utilizing a weakly-supervised approach specialized for alignment, we follow the approach of \cite{dwibedi2019temporal} and train a separate model for each of the 13 action classes in this dataset. The results shown in the table are the average across all 13 actions; for a more detailed breakdown of results see Appendix~\ref{app: PennAction Breakdown}, Table~\ref{tab: PennAction breakdown}. 

Our method consistently outperforms all prior work, as highlighted in bold, with SCL performing the second best overall. It is important to note, however, that SCL trained a single model for all 13 action classes (the authors did not report results for each class individually), whereas our method utilizes individualized models for each class. Furthermore, our method achieves an accuracy of $93.17\%$ using only 75\% of available labels, surpassing all baselines in Phase Classification (even if they use all the labels).
Additionally, we achieve more than a 1\% improvement in the Phase Progression metric. This suggests that our method produces highly reliable features for video alignment and fine-grained retrieval, as supported by the small but consistent improvement in Kendall's tau and Average Precision@K metrics. As we did not re-train SCL on this dataset, we do not report the DTW Accuracy in the table. {\bf LRProp} achieves an impressive average DTW Accuracy of $90.15$\%.

\subsection{Ablation Study}

SoftDTW and pair-wise position propagation: Is it a winning combination? We now turn to address this question by assessing the performance of these two loss terms, $\mathcal{L}_{\text{Prop.}}$ and $\mathcal{L}_{\text{Sdtw}}$, and their contribution on the pouring dataset. We measure the Phase Classification Accuracy for a more diverse set of label portions, and similarly, we measure the Average Precision accuracy for a larger range of K values. We also measure the Phase Progression, Kendall's Tau and DTW Accuracy as in previous sections. The results are depicted in Table~\ref{tab: ablation}.

As can be seen, our design choices consistently improve across all evaluation metrics, with the exception of Kendall's Tau, which is already close to saturation. We observe that the first and second rows of Table~\ref{tab: ablation} have similar results in the Kendall's Tau, Phase Progression, and Average Precision@K metrics. This suggests that the combination of $\mathcal{L}_{\text{Prop.}}$ and $\mathcal{L}_{\text{Sdtw}}$ is responsible for the improvement, rather than either one alone. Additionally, \textbf{LRProp} shows an average improvement of around 5\% in the Phase Classification Accuracy metric for any choice of label percentage. Also, the striking improvement in the DTW Accuracy metric when using both $\mathcal{L}_{\text{Prop.}}$ and $\mathcal{L}_{\text{Sdtw}}$ is particularly noteworthy, as it suggests that the combined loss function is able to produce features that are highly effective for video alignment.

% \usepackage{multirow}

% \usepackage{multirow}
% \usepackage[normalem]{ulem}
% learning representations by position propagation
\section{Conclusions}
In this paper we introduce \textbf{LRProp}, a powerful method for extracting frame-level features that are particularly effective for aligning videos. Our approach involves using the dynamic time-warping (DTW) algorithm for establishing a prior distribution between frames from different videos based on their alignment path. In multiple experiments and commonly used metrics, we demonstrate that our method significantly outperforms various baselines in learning feature representations. We also evaluate DTW video alignment performance using the learned features and achieve a striking improvement of more than 5\% compared to the previous state-of-the-art.

A potential area for future research is to adapt our method to work with videos that may contain outlier frames, by identifying and extracting them from the learning/inference processes. This may enable more robust features and better overall alignment between videos. Another challenging task refers to much longer videos, for which current solutions may face a severe memory and computational barriers. 
% Additionally, as the alignment path is not a one-to-one mapping between frames, our current technique for \emph{pair-wise position propagation} involves selecting the smallest index when multiple indices are connected to the same frame. However, it could be interesting to explore more advanced methods, such as taking the average of all connections or using KNN; or maybe even soft nearest neighbors, to learn the most plausible mapping between frames. 

To conclude, our results demonstrate that the proposed approach is highly effective at learning better frame-wise feature representations for videos. The use of \emph{pair-wise position propagation}, combined with the SoftDTW, to relate frames between different videos is a particularly promising direction, as we demonstrate its potential to significantly improve the task of video alignment.

%%%%%%%%%%%%%%%%%%%%%%%%%%%%%%%%%%%%%%%%%%%%%%%%%%%%%%%%%%%%
\newpage
% \nocite{*}
\bibliography{main}
\bibliographystyle{vancouver}

\appendix
\onecolumn

\section{Appendix}

% \subsection{Alignment Visualization}
% \label{app: Alignment Visualization}
% In this section, we present a visualization of the alignment of videos from the pouring dataset using the DTW algorithm, on the features extracted by \textbf{LRProp}.
% We showcase the key stages in the pouring action, including lifting the bottle off, starting to pour, completing the pour, and setting the bottle down. In Figure~\ref{fig: alignment}, we have selected a random video from the pouring dataset (first row) and aligned it using DTW and our features with five additional randomly chosen videos (next five rows). The aligned frames corresponding to the key events in the randomly selected video are displayed. It is evident that \textbf{LRProp} effectively captures the key events in the query video.

% \begin{figure*}[h]
    
%     \centering
%     \includegraphics[width=\textwidth]{figures/method/aligment_vis.png}
%     \caption{\textbf{Pouring.} DTW alignment demonstration.}
%     \label{fig: alignment}
% \end{figure*}

\subsection{PennAction Breakdown}
\label{app: PennAction Breakdown}

In this section, we present a detailed breakdown of \textbf{LRProp} performance on each action in the PennAction dataset. The results are summarized in Table~\ref{tab: PennAction breakdown}.
% \usepackage{multirow}
\begin{table*}[h]
\tiny
\begin{tabular}{l||l|l|lllllll|llllll|l}
\multicolumn{1}{c||}{\multirow{2}{*}{\textit{Dataset}}} & \multicolumn{1}{c|}{\multirow{2}{*}{\textit{$\tau$}}} & \multicolumn{1}{c|}{\multirow{2}{*}{\textit{Progress}}} &       & \multicolumn{6}{c|}{\textit{AP@}}                                                                                                          & \multicolumn{6}{c|}{\textit{Classification@}}                                                                                                                & \multicolumn{1}{c}{\multirow{2}{*}{\textit{DTW-A}}} \\
\multicolumn{1}{c||}{}                                  & \multicolumn{1}{c|}{}                                                  & \multicolumn{1}{c|}{}                                   & K=1   & \multicolumn{1}{c}{\textit{K=5}} & \multicolumn{1}{c}{\textit{K=10}} & \multicolumn{1}{c}{\textit{K=15}} & K=20  & K=25  & K=30  & 5     & \multicolumn{1}{c}{\textit{10}} & \multicolumn{1}{c}{\textit{25}} & \multicolumn{1}{c}{\textit{50}} & 75    & \multicolumn{1}{c|}{\textit{100}} & \multicolumn{1}{c}{}                       \\ \midrule \midrule
squat                                                 & 99.58                                                                 & 97.08                                                  & 90.9  & 90.73                            & 90.22                             & 90.11                             & 89.95 & 89.83 & 89.74 & 91.19 & 90.43                           & 91                              & 91.07                           & 91    & 91.57                            & 87.4                                       \\
pushup                                                & 99.08                                                                 & 92.7                                                   & 92.58 & 92.53                            & 92.39                             & 92.32                             & 92.2  & 92.12 & 92.02 & 90.77 & 89.27                           & 91.86                           & 93.17                           & 93.25 & 93.19                            & 92.35                                      \\
tennis serve                                          & 98.66                                                                 & 95.54                                                  & 90.05 & 89.44                            & 89.2                              & 89.03                             & 88.76 & 88.58 & 88.47 & 89.4  & 91.31                           & 91.52                           & 91.17                           & 91.74 & 91.69                            & 87.85                                      \\
tennis forhand                                        & 98.85                                                                 & 88.28                                                  & 92.14 & 91.03                            & 90.89                             & 90.61                             & 90.47 & 90.36 & 90.3  & 92.53 & 92.44                           & 93.14                           & 92.68                           & 93.14 & 93.08                            & 89.66                                      \\
situp                                                 & 99.33                                                                 & 95.39                                                  & 96.83 & 96.81                            & 96.7                              & 96.63                             & 96.51 & 96.41 & 96.4  & 96.81 & 96.72                           & 97.04                           & 96.97                           & 97.08 & 97.08                            & 95.74                                      \\
pullup                                                & 99.23                                                                 & 96.8                                                   & 96.73 & 96.52                            & 96.46                             & 96.39                             & 96.33 & 96.23 & 96.18 & 96.4  & 96.34                           & 95.66                           & 95.54                           & 96.13 & 96.27                            & 95.25                                      \\
jumping jacks                                         & 98.63                                                                 & 97.02                                                  & 91.91 & 91.84                            & 91.74                             & 91.67                             & 91.44 & 91.03 & 90.68 & 92.11 & 93.41                           & 93.58                           & 93.58                           & 93.7  & 93.76                            & 90.08                                      \\
golf swing                                            & 98.92                                                                 & 98.19                                                  & 95.02 & 95.3                             & 94.98                             & 94.64                             & 94.46 & 94.26 & 94.14 & 94.92 & 94.92                           & 94.97                           & 95.28                           & 95.37 & 95.4                             & 92.69                                      \\
bowl                                                  & 98.84                                                                 & 76.21                                                  & 92.33 & 90.65                            & 89.73                             & 89.13                             & 88.56 & 88.22 & 87.85 & 84.57 & 85.02                           & 84.53                           & 84.9                            & 85.25 & 85.68                            & 83.88                                      \\
benchpress                                            & 99.32                                                                 & 95.14                                                  & 94.33 & 93.96                            & 93.9                              & 93.75                             & 93.65 & 93.55 & 93.49 & 91.38 & 93.81                           & 94.98                           & 95.11                           & 95.19 & 95.33                            & 92.57                                      \\
baseball swing                                        & 99.2                                                                  & 94.9                                                   & 91.61 & 92.11                            & 91.87                             & 91.88                             & 91.78 & 91.63 & 91.49 & 97.05 & 89.59                           & 92.49                           & 93.03                           & 93.19 & 93.11                            & 90.07                                      \\
baseball pitch                                        & 99.35                                                                 & 98.05                                                  & 92.92 & 92.91                            & 92.73                             & 92.59                             & 92.55 & 92.43 & 92.32 & 92.18 & 93.11                           & 94.62                           & 94.94                           & 95.18 & 95.29                            & 90.93                                      \\
clean and jerk                                        & 99.2                                                                  & 87.64                                                  & 88.67 & 88.24                            & 87.84                             & 87.66                             & 87.5  & 87.33 & 87.2  & 87.89 & 88.62                           & 90.31                           & 91.14                           & 90.99 & 90.8                             & 83.59   \\ \midrule  \midrule                                  
\end{tabular}

\caption{\textbf{PennAction breakdown}. A breakdown of the performance of \textbf{LRProp} on each of the actions contained in the PennAction dataset, using various evaluation metrics: Phase Classification@\% (Classification@\%),
Phase Progression (Progress), Kendall’s Tau ($\tau$), Average Precision@K (AP@K) and DTW Accuracy (DTW A).}
\label{tab: PennAction breakdown}
\end{table*}

\subsection{Implementation Details}
\label{app: Implementation Details}
In our model, we use a ResNet-50~\cite{he2016deep} pretrained by BYOL~\cite{grill2020bootstrap} as a frame-wise spatial encoder. We use a 7-layer Transformer encoder~\cite{vaswani2017attention} with a hidden size of 256 and 8 heads to model temporal context. We set $\sigma^2=10$, recall Equations~(\ref{eq: prior for same video}),~(\ref{eq: prior for different video}), and $\tau=0.1$, recall Equation~(\ref{eq: embedding similarity}). We use the Adam optimizer with a learning rate of $10^{-4}$ and a weight decay of $10^{-5}$, and we apply a cosine decay schedule without restarts~\cite{loshchilov2016sgdr} to the learning rate. We fix a batch size of 2, and train for 300 epochs all considered datasets. We use the same spatial augmentations and temporal sampling technique as in~\cite{chen2020simple, chen2022frame}. The number of sampled frames, $T$, and the regularization parameters, $\lambda_1, \lambda_2$, recall Equation~(\ref{eq: final loss}), are given in Table~\ref{tab: hyperparameters}. 

% In our model, we adopt ResNet-50~\cite{xxx}, pretrained by BYOL, as a frame-wise spatial encoder. We use a 7-layer Transformer encoder~\cite{xxx}, with 256 hidden size and 8 heads to model temporal context. We train the model using Addam optimizer with learning rate $10^{-4}$ and weight decay $10^{-5}$. We decay the learing rate with cosine decay schedule without restarts~\cite{xxx}. We set $\sigma^2=10$, recall Equation~\ref{xxx}, and $tau=0.1$, recall Equation~\ref{xxx}. Following~\cite{xxx, xxx}, we deploy the same spatial augmentations and temporal sampling technique, for the actual number of sampled frames, $T$, regarding each dataset, see Table~\ref{xxx}. We use a batch size of 2 for all datasets, and our model is trained for 300 epochs. for the values of $\lambda_1, \lambda_2$, recall Equation~\ref{xxx}, also see Table~\ref{tab: hyperparameters}.

% \usepackage{multirow}
\begin{table}[h]
\centering
\begin{tabular}{c||c|c|c|c}
\multirow{2}{*}{\textit{Dataset}} & \multirow{2}{*}{\textit{Action class}} & \multirow{2}{*}{\textit{T}} & \multirow{2}{*}{$\lambda_1$} & \multirow{2}{*}{$\lambda_2$} \\
                                  &                                        &                             &                              &                              \\ \midrule \midrule
\multirow{13}{*}{PennAction}      & jumping jacks                          & 20                          & \multirow{14}{*}{0.01}       & 0.4                          \\ \cline{2-3} \cline{5-5} 
                                  & tennis serve                           & \multirow{3}{*}{40}         &                              & \multirow{2}{*}{0.003}       \\ \cline{2-2}
                                  & tennis forhand                         &                             &                              &                              \\ \cline{2-2} \cline{5-5} 
                                  & baseball swing                         &                             &                              & \multirow{2}{*}{0.001}       \\ \cline{2-3}
                                  & bowl                                   & 60                          &                              &                              \\ \cline{2-3} \cline{5-5} 
                                  & situp                                  & \multirow{8}{*}{80}         &                              & \multirow{3}{*}{0.4}         \\ \cline{2-2}
                                  & benchpress                             &                             &                              &                              \\ \cline{2-2}
                                  & baseball pitch                         &                             &                              &                              \\ \cline{2-2} \cline{5-5} 
                                  & golf swing                             &                             &                              & \multirow{2}{*}{1.6}         \\ \cline{2-2}
                                  & pushup                                 &                             &                              &                              \\ \cline{2-2} \cline{5-5} 
                                  & squat                                  &                             &                              & \multirow{4}{*}{0.8}         \\ \cline{2-2}
                                  & pullup                                 &                             &                              &                              \\ \cline{2-2}
                                  & clean and jerk                         &                             &                              &                              \\ \cline{1-3}
Pouring                           & -                                      & 240                         &                              &                              \\ \midrule \midrule
\end{tabular}
\caption{Hyper-parameters of \textbf{LRProp} regarding each dataset.}
\label{tab: hyperparameters}
\end{table}

\end{document}